\documentclass[pdflatex,sn-basic,iicol]{sn-jnl}

\usepackage{graphicx}
\usepackage{multirow}
\usepackage{amsmath,amssymb,amsfonts}
\usepackage{amsthm}
\usepackage{mathrsfs}
\usepackage[title]{appendix}
\usepackage{xcolor}
\usepackage{textcomp}
\usepackage{manyfoot}
\usepackage{booktabs}
\usepackage{arydshln}
\usepackage{algorithm}
\usepackage{algorithmicx}
\usepackage{algpseudocode}
\usepackage{listings}

\DeclareMathOperator*{\argmax}{arg\,max}

\theoremstyle{thmstyleone}

\theoremstyle{thmstyletwo}

\theoremstyle{thmstylethree}

\begin{document}

\title[Article Title]{Corner-Grasp: Multi-Action Grasp Detection and Active Gripper Adaptation for Grasping in Cluttered Environments}

\author[1]{\fnm{Yeong Gwang } \sur{Son}}\email{syoungk20@g.skku.edu}

\author[1]{\fnm{Seunghwan } \sur{Um}}\email{rush@g.skku.edu}

\author[1]{\fnm{Juyong } \sur{Hong}}\email{juyong0000@skku.edu}

\author*[1]{\fnm{Hyouk Ryeol } \sur{Choi}}\email{choihyoukryeol@gmail.com}

\affil[1]{\orgdiv{Department of Mechanical Engineering}, \orgname{Sungkyunkwan University}, \orgaddress{\street{Suwon}, \city{Gyeonggi-do}, \postcode{16419}, \state{South Korea}, \country{Korea}}}

\abstract{Robotic grasping is an essential capability, playing a critical role in enabling robots to physically interact with their surroundings. Despite extensive research, challenges remain due to the diverse shapes and properties of target objects, inaccuracies in sensing, and potential collisions with the environment. In this work, we propose a method for effectively grasping in cluttered bin-picking environments where these challenges intersect. We utilize a multi-functional gripper that combines both suction and finger grasping to handle a wide range of objects. We also present an active gripper adaptation strategy to minimize collisions between the gripper hardware and the surrounding environment by actively leveraging the reciprocating suction cup and reconfigurable finger motion. To fully utilize the gripper's capabilities, we built a neural network that detects suction and finger grasp points from a single input RGB-D image. This network is trained using a larger-scale synthetic dataset generated from simulation. In addition to this, we propose an efficient approach to constructing a real-world dataset that facilitates grasp point detection on various objects with diverse characteristics. Experiment results show that the proposed method can grasp objects in cluttered bin-picking scenarios and prevent collisions with environmental constraints such as a corner of the bin.}

\keywords{Picking in clutter, Bin-picking, Multi-functional gripper, Computer vision for automation, Dataset for robotic grasping}

\maketitle

\section{Introduction}\label{sec1}

Robotic grasping is a key enabler for automation in logistics, manufacturing, and service environments, where robots must reliably pick, transport, and place objects under real-world uncertainty. Among these applications, cluttered bin-picking remains particularly challenging. This difficulty arises because target objects are often partially occluded, densely stacked, and positioned close to environmental constraints such as bin walls and corners. In these specific settings, a grasping system must meet two critical requirements. It needs to generalize to previously unseen objects with diverse geometries and materials. At the same time, it must execute grasps without colliding with the surrounding environment. These conditions frequently co-occur in practical warehouse workflows, making the task even more demanding.

A central difficulty arises from the diversity of graspable object properties. Suction-based grasping is attractive for fast picking and reduced risk of damaging objects, but its effectiveness depends on forming a seal on the target surface. Because of this requirement, the method frequently fails on porous, cloth-like, highly curved, or irregular surfaces. In contrast, finger-based grasping can handle a broader range of shapes and achieve secure form or force closure when contact is feasible. However, this approach is highly sensitive to available clearance. Objects that are too large, too tightly packed, or too close to walls may prevent fingers from approaching and closing safely. As a result, single-modality systems often exhibit brittle behavior when deployed across heterogeneous object sets, especially under heavy clutter.

\begin{figure}[t!]
    \centering
    \includegraphics[width=\columnwidth]{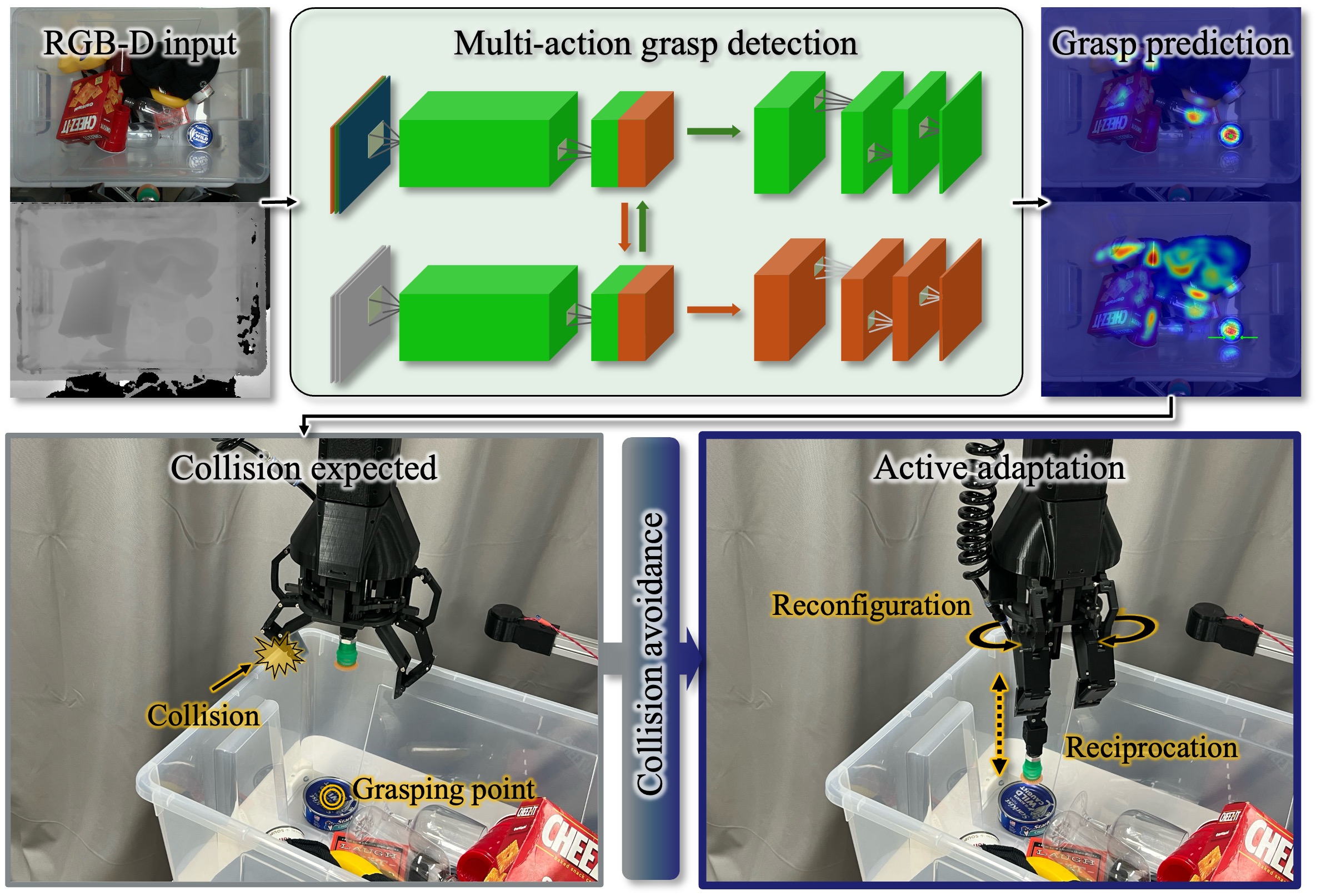}
    \caption{Overview of the proposed system. The system detects multi-action grasp candidates from a single RGB-D image. It also incorporates an active collision-avoidance strategy that utilizes the gripper’s reconfigurable finger motion and reciprocating suction cup to avoid collisions with environmental constraints, enabling reliable grasping even in constrained conditions such as bin corners}
    \label{fig:overview}
\end{figure}

One direct approach is to build a robot equipped with multiple end-effectors, such as independent suction and finger grippers, and select the appropriate tool for each object. While this improves coverage, it typically increases hardware cost, volume, and integration complexity. Additionally, these multi-tool setups can be difficult to deploy in space-constrained industrial cells. This limitation has motivated growing interest in multi-functional hybrid grippers that integrate both suction and finger grasping within a single end-effector. Such grippers offer the promise of covering complementary failure modes while maintaining a compact form factor. However, effectively exploiting multi-functional hardware in clutter is not automatic. It requires advanced perception and control systems. Specifically, the system must predict multiple grasp modes, select among them robustly, and adapt the gripper configuration and motion to avoid collisions in confined workspaces.

Existing learning-based grasp detection has made substantial progress toward generalization to unseen objects, especially with model-free approaches trained on large-scale data~\cite{kleeberger2020survey}. Yet, many methods still predict grasps for a single modality, such as using only suction~\cite{son2024coasnet, jiang2023multiple, schillinger2023model} or only fingers~\cite{bui2023antipodal, hieu2024enhancing, gou2021rgb}. These methods implicitly assume that a successfully predicted grasp can be executed exactly as planned. In real bins, this assumption frequently breaks down. Even when a highly confident grasp point exists, the robot may be unable to reach it due to bin walls, corner geometry, or interference between the end-effector hardware and neighboring objects. In practice, many experimental evaluations avoid these issues by simplifying the environment with shallow bins or flat tabletop setups~\cite{cao2021suction, fang2020graspnet}. This simplification obscures a major source of failures encountered during actual deployment.

This paper addresses the intersection of these challenges by proposing a method for the reliable grasping of diverse and previously unseen objects in cluttered bins. The approach explicitly accounts for execution feasibility under strong environmental constraints, such as bin corners, using a compact multi-functional gripper. To achieve this goal, we introduce Corner-Grasp. This grasping system couples multi-action grasp detection with active gripper adaptation to ensure collision-aware execution in constrained bin environments. Our system uses a multi-functional gripper that integrates suction and finger grasping. It further leverages a reciprocating suction cup and reconfigurable finger motion as actionable degrees of freedom. These hardware characteristics allow for effective collision mitigation near bin walls and corners. On the perception side, we designed a unified neural network that predicts both suction and finger grasp candidates from a single RGB-D observation. This design enables fast and consistent multi-modal inference without the need to run separate pipelines. Robust multi-action learning demands scale and coverage. Therefore, we trained the network using large-scale photorealistic synthetic data. We then complemented this training with a real-world dataset constructed with minimal manual effort through efficient label propagation across controlled viewpoint variations. Finally, we incorporated a lightweight material-aware module. This module triggers auxiliary strategies for object categories that frequently cause failures in practice, such as transparent or metallic items. An overview of the proposed system is illustrated in Fig.~\ref{fig:overview}.

Extensive real-robot experiments in cluttered scenarios demonstrate that the proposed approach grasps a wide range of object types and significantly reduces collision-related failures in deep bin settings, including corner cases where conventional pipelines often fail. Beyond bin clearing performance, we also validate that the system generalizes across variations in backgrounds, bin configurations, and object sets, supporting practical deployment conditions.

Our contribution can be summarized as follows:
\begin{itemize}
    \item We introduce a unified grasp detection network that simultaneously estimates suction and finger grasp affordances from a single RGB-D input, enabling efficient selection between grasp modes for diverse objects in clutter.
    \item We propose a collision-aware execution strategy that actively leverages the gripper’s reciprocating suction cup and reconfigurable fingers to mitigate collisions near bin edges or corners, including adaptive reconfiguration and interaction behaviors when direct execution is infeasible.
    \item We build a large-scale photorealistic synthetic dataset with dense multi-action grasp annotations and propose an efficient real-world data collection/labeling procedure that substantially reduces human annotation cost while improving real-world robustness.

\end{itemize}

Multimedia material accompanying this paper is provided in the supplementary material including video demonstrations of the proposed system.

\section{Related Work}\label{sec2}

\subsection{Gripper Design}\label{subsec1}
Gripper design fundamentally determines a robotic system's grasping capabilities. Finger grippers, including two-fingered~\cite{yoon2022fully, sun2022larg} and multi-fingered~\cite{li2015reconfigurable, liu2020optimal, ma2017yale} variants, adapt to diverse shapes and can enhance force via adaptive mechanisms~\cite{issac2024gripper, Kobayashi2019grip}. However, they require sufficient collision-free space for finger placement and risk damaging fragile objects. Conversely, suction grippers~\cite{eppner2018four, schwarz2017data} offer damage-free manipulation but strictly require flat, non-porous surfaces to maintain vacuum seals. To overcome these isolated limitations, multi-functional grippers integrating several modalities and additional degrees of freedom (DoF) have emerged~\cite{zeng2022robotic, fang2020softgrip, nagata2021modeling, salvatore2023planner}. Despite these hardware advancements, prior research predominantly evaluates grasping in unconstrained table-top setups, treating collisions as a secondary concern. In real-world bin-picking, spatial constraints such as walls and corners make gripper protrusions, opening space, and approach angles critical failure factors.

Our work builds upon multi-functional hardware, specifically leveraging the gripper's DoF as an explicit tool for collision-aware execution to ensure effectiveness in tightly constrained bin environments.

\subsection{Robotic Grasping in Cluttered Environments}\label{subsec2}
Recent data-driven methods utilizing CNNs and transformers have significantly advanced robotic grasping by generalizing to unseen objects across varying environments. Grasp poses are typically parameterized as either top-down (position and z-axis rotation) or 6D (full 3D pose). While extensive literature focuses on single-modality detection predicting either finger~\cite{bui2023antipodal, hieu2024enhancing, chen2023efficient, fang2020graspnet, kumra2020antipodal, mahler2017dex, gou2021rgb} or suction grasps~\cite{cao2021suction, son2024coasnet, mahler2018dex, jiang2023multiple,schillinger2023model}, these approaches often fail systematically on heterogeneous object sets due to inherent end-effector limitations. To overcome this, multi-modal frameworks~\cite{zeng2022robotic, liu2023hybrid} predict and select from multiple grasp types. However, these methods are rarely evaluated under strong environmental constraints, leaving execution feasibility largely underexplored~\cite{mahler2019dexnet4, salvatore2024cepb, hasegawa2019graspfusion}.

Our approach addresses this limitation by integrating multi-action grasp detection (suction and fingers) with a smart execution layer that explicitly reasons about constrained bin regions. By actively leveraging gripper degrees of freedom—such as reciprocation and reconfiguration—to reduce collision risks, we aim to bridge the gap between visual grasp detection and reliable physical execution in deep, cluttered environments.

\begin{figure}[t!]
    \centering
    \includegraphics[width=\columnwidth]{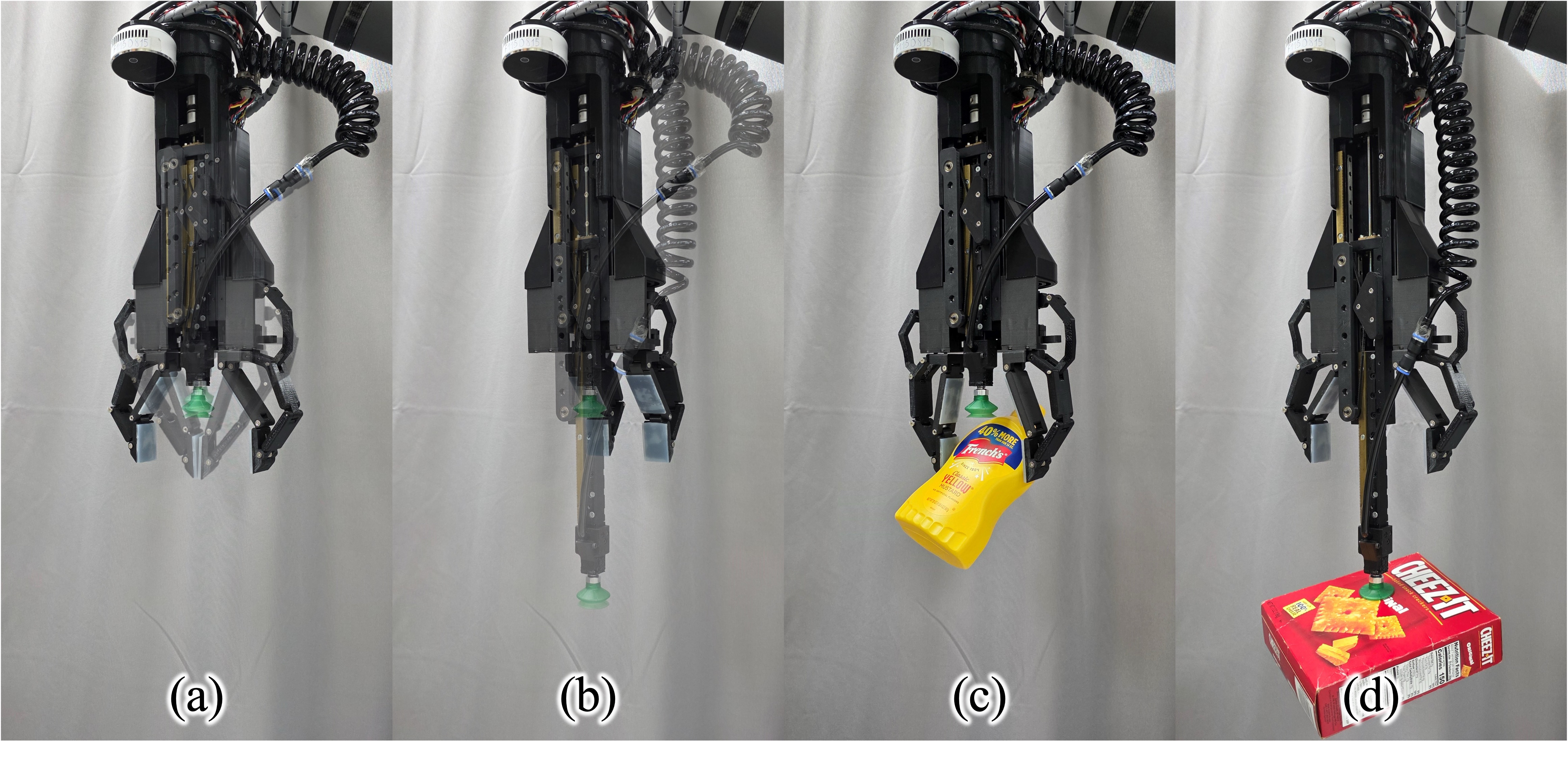}
    \caption{Functionalities of ReC-Gripper based on its configurations: (a) Finger grasping motion; (b) Reciprocating suction motion;
     (c) Grasping an object using finger grasp mode; (d) Grasping an object using suction grasp mode}
    \label{fig:gripper_modes}
\end{figure}

\subsection{Dataset for Robotic Grasping}\label{subsec3}

Data-driven grasp detection relies heavily on large-scale datasets. Traditionally, real-world data collection required labor-intensive manual annotation~\cite{depierre2018jacquard, zeng2022robotic}. To mitigate this, photorealistic simulators now automatically generate vast datasets with precise annotations using physics-based evaluations~\cite{9560844}, analytical models leveraging privileged information~\cite{son2024coasnet, hieu2024enhancing, mahler2018dex, mahler2017dex}, or hybrid approaches~\cite{MetaGraspNetV2, MetaGraspNet}. However, the simulation-to-reality gap, often caused by difficult-to-model physical properties and sensor artifacts from transparent objects, necessitates the inclusion of real-world data, which reintroduces the manual annotation bottleneck.

In this study, we introduce an analytically annotated, synthetic bin-picking dataset generated via Isaac Sim, extending our previous work~\cite{son2024coasnet, hieu2024enhancing} without requiring human labeling. To effectively bridge the sim-to-real gap and optimize practical performance, we additionally propose an efficient method for constructing a complementary real-world dataset with minimal human intervention.

\section{Problem Statement}\label{sec3}

Let $G=\{T, C\}$ be a grasp candidate defined by a grasp pose $T=\{R, t\}\in \mathbb{S}\mathbb{E}(3)$ and a gripper configuration $C=(j_{s}, j_{f}, j_{r})$, where $R$ and $t$ denote the grasp rotation matrix and position vector, respectively, $j_{s}$ is the suction joint state, $j_{f}$ is the finger joint state, and $j_{r}$ is the finger rotation joint state. The goal is to detect grasp candidates from RGB and depth images, $I_{rgb}\in \mathbb{R}^{3\times H\times W}, I_{d}\in \mathbb{R}^{H\times W}$ captured by an overhead camera mounted on the wrist of the manipulator. The system should be able to detect either suction or finger grasp candidates and select the best grasp candidate that maximizes the grasp success probability despite uncertainty caused by noisy input. Additionally, the system should handle various object shapes, sizes, and materials and adapt to novel objects that were not seen during training. Furthermore, it should avoid collisions with the environment and other objects in the bin.

\section{Gripper Design}\label{sec:gripper_design}

In this section, we introduce a multi-functional gripper that combines both suction and finger grasping to handle a wide range of objects in the bin-picking task. We also present an auxiliary electromagnetic grasping strategy that can be applied to the proposed system without modifying the gripper hardware.

\subsection{Multi-functional Gripper}

\begin{figure}[t!]
    \centering
    \includegraphics[width=\columnwidth]{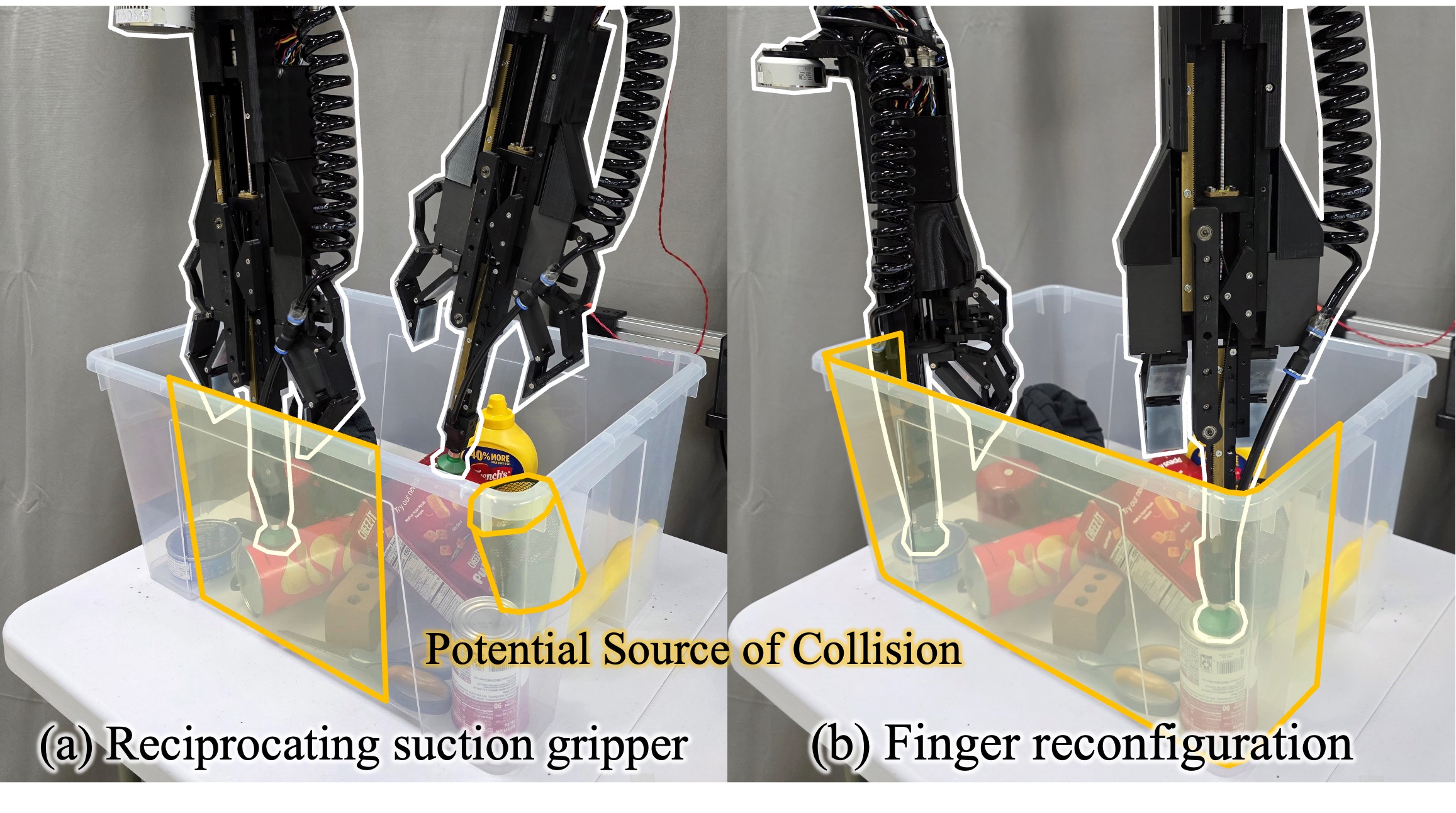}
    \caption{Adaptations of ReC-Gripper for grasping in cluttered environments: (a) Reciprocating suction cup motion to avoid potential collision sources, such as bin walls or objects near the grasping point; (b) Reconfigurable finger motion to prevent collisions when grasping in environmental constraints such as corner spaces}
    \label{fig:gripper_adaptations}
\end{figure}

ReC-Gripper~\cite{um2024rec} is a reconfigurable integrated suction and fingered gripper characterized by a zeroed offset between the suction gripper and the finger, finger reconfiguration, and a suction cup that has a reciprocating motion. This gripper design enables a supporting motion with its fingers to hold up an object in shelf-picking scenarios demonstrating superior performance in stable manipulation under shelf-pick and stowing tasks. The gripper compensates for the limitations of single-function grippers by utilizing various grasping modes, enabling it to handle a wide range of objects that are challenging for conventional single-mode grippers. Fig.~\ref{fig:gripper_modes} illustrates the various modes of ReC-Gripper, as well as the reciprocating motion of the suction gripper and the reconfiguration of the fingers. The state of the suction gripper, denoted as $j_s$, is categorized into an extended state $j_s=s_{out}$ and a retracted state $j_s=s_{in}$. The state of the fingers, denoted as $j_f$, consists of a closed state $j_f=f_{close}$, an open state $j_f=f_{open}$, 0-degree rotated default state $j_r=f_{dft}$, and a 90-degree rotated state $j_r=f_{rot}$.

\subsection{Adaptations to Bin Picking Task}

\begin{figure}[t]
    \centering
    \includegraphics[width=\columnwidth]{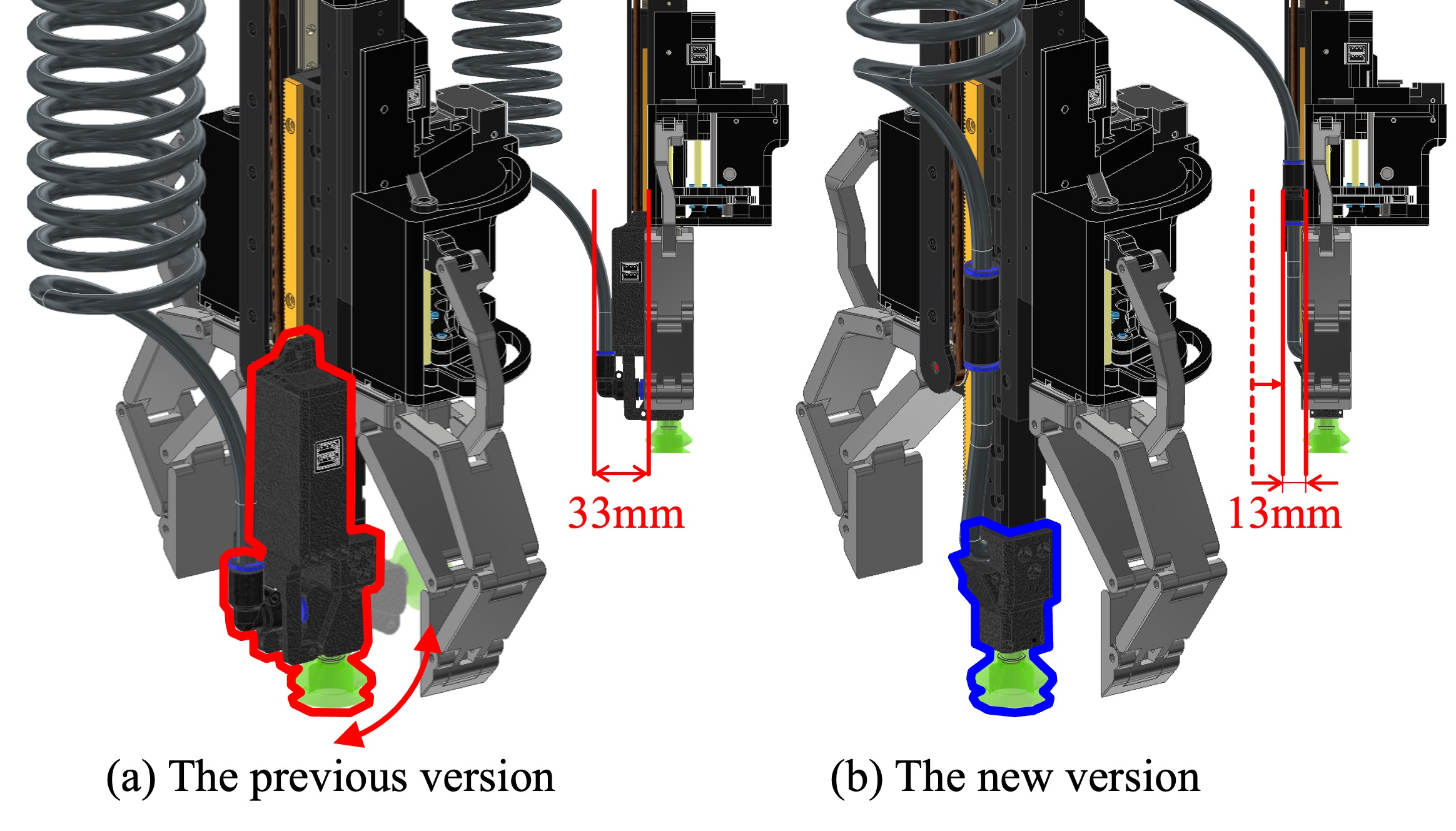}
    \caption{The gripper modification: (a) Previous gripper design with protrusions that can cause collisions; (b) Modified gripper design with a flat side to minimize potential collisions}
    \label{fig:gripper_modification}
\end{figure}

We found that the design of ReC-Gripper can be adapted to the bin-picking task by leveraging its functionalities. The reciprocating suction cup can be used to avoid collisions with the bin walls and other objects in the bin. Additionally, the reconfigurable finger motion, which is designed to support objects in shelf-picking tasks, can be utilized to lessen collisions during grasping. Fig.~\ref{fig:gripper_adaptations} illustrates the gripper's adaptations to the bin-picking task. To further reduce the risk of collisions and improve workspace utilization, we modified the gripper hardware to make one side flat without any protrusions that could cause collisions. This modification allows the gripper to grasp objects near the bin walls without colliding with them. Fig.~\ref{fig:gripper_modification} shows the modified gripper design.

\begin{figure}[t]
    \centering
    \includegraphics[width=\columnwidth]{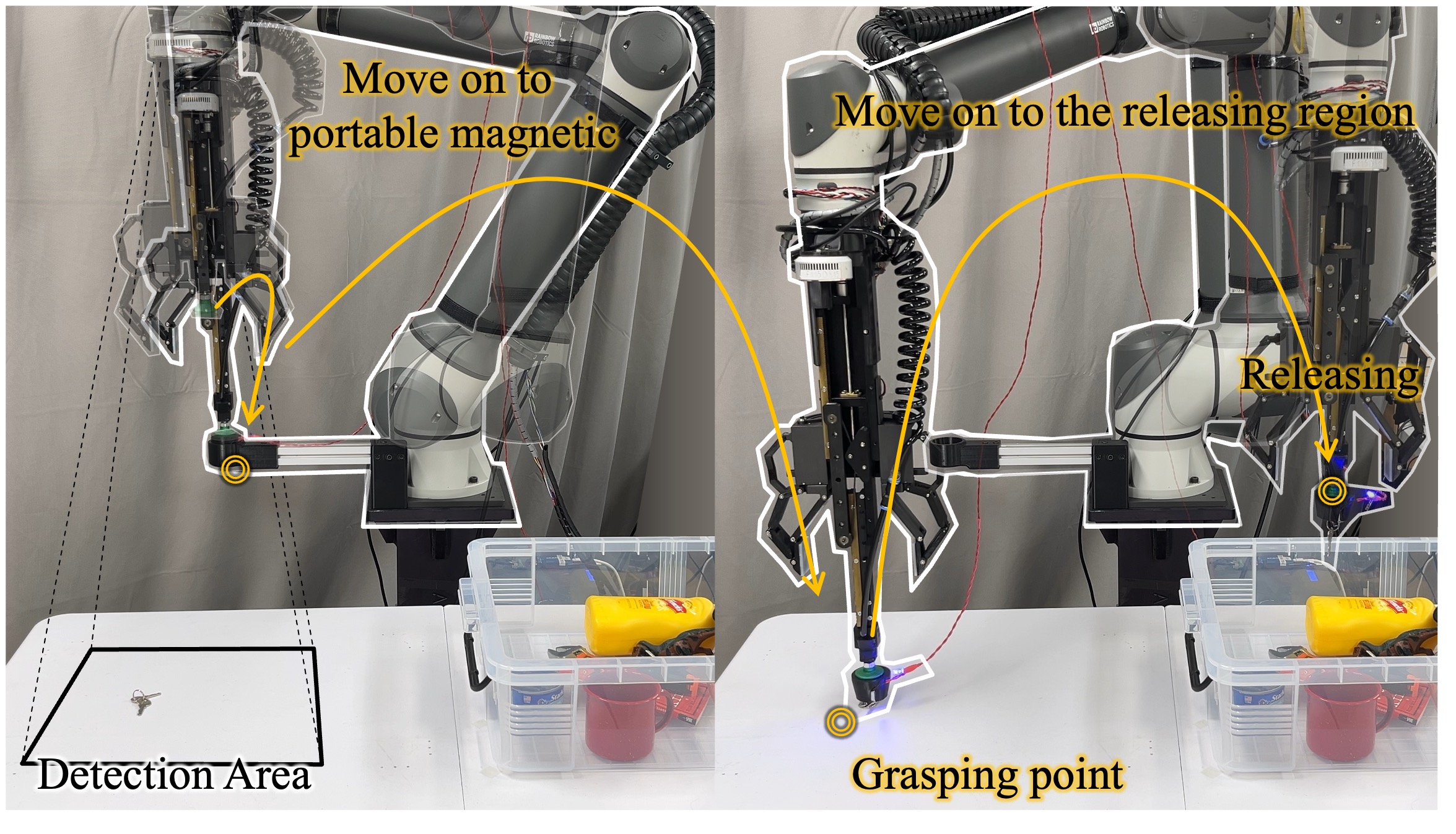}
    \caption{The electromagnetic grasping strategy of our system. When the system detects a metallic object, the gripper picks up the electromagnetic holder placed near the base of the robot arm using the suction cup and moves it to the target grasping point. The electromagnetic holder then attracts the metallic object, allowing the gripper to pick it up}
    \label{fig:magnetic_grasping}
\end{figure}

\begin{figure*}[th!]
    \centering
    \includegraphics[width=\textwidth]{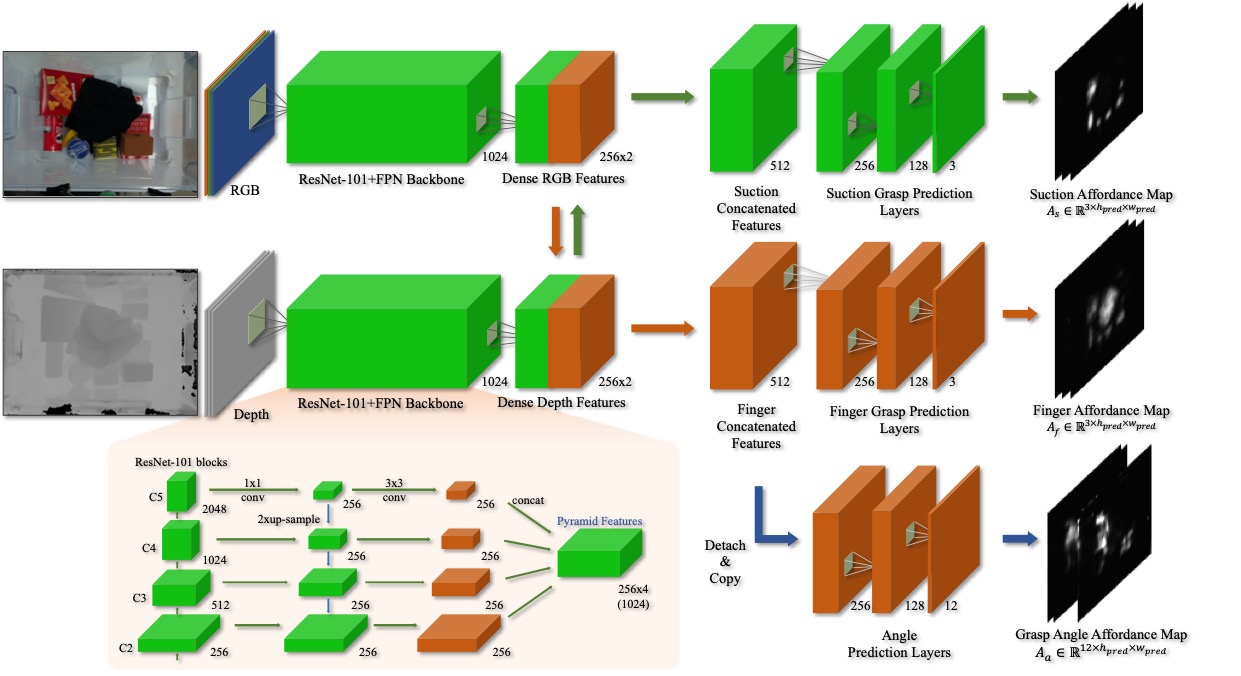}
    \caption{The proposed multi-action grasp detection network architecture. The network consists of RGB and depth trunks for shared feature extraction. Task-specific output heads predict the suction affordance map, finger affordance map, and grasp angle affordance map}
    \label{fig:multi_action_grasp_network}
\end{figure*}
\subsection{Electromagnetic Grasping}
Although the multi-functional gripper can handle a variety of object types, grasping tiny metallic objects remains challenging, as is the case in most previous research. Grasping tiny metallic objects is difficult primarily due to the limited contact area for grasping that forms either force or form closure. Instead of modifying the gripper hardware to handle metallic objects, we propose an auxiliary electromagnetic grasping strategy that can be applied to the proposed system.
To implement the electromagnetic grasping strategy, we positioned an electromagnetic holder near the base of the robot arm. When the system detects a metallic object, the gripper picks up the electromagnetic holder using the suction cup and moves it to the target object. The electromagnetic holder then attracts the metallic object, allowing the gripper to pick it up. This strategy enables the system to grasp metallic objects without modifying the gripper hardware. Fig.~\ref{fig:magnetic_grasping} illustrates the electromagnetic grasping strategy.

\section{Multi-Action Grasp Detection}

In this section, we propose a multi-action grasp detection model capable of predicting suction and finger grasp points end-to-end to leverage the diverse functionalities of the gripper. We also introduce a surface material recognition method to enhance the system's ability to handle diverse object types. Additionally, we present a collision avoidance planning strategy that actively utilizes the gripper functionalities to prevent environmental collisions. Finally, we propose a grasp fusion method to improve grasping stability.

\subsection{Grasp Detection Model}

Our previously developed suction grasp detection network and finger grasp detection network~\cite{son2024coasnet,hieu2024enhancing} have demonstrated their effectiveness in grasping a wide range of unseen objects in real-world environments. Building upon this study, we propose a multi-action grasp detection network capable of performing both suction and finger grasping. Rather than independently inferring results using separate networks, we designed a unified network architecture that integrates both functionalities. To achieve this, we employed a multi-task learning (MTL) approach based on a shared trunk architecture. As illustrated in Fig.~\ref{fig:multi_action_grasp_network}, the proposed network utilizes RGB and depth trunks composed of ImageNet-pretrained ResNet-101~\cite{he2016resnet} and Feature Pyramid Network (FPN)~\cite{lin2017fpn} to extract shared features from input scenes. Subsequently, task-specific output heads are deployed to predict the suction affordance map, finger affordance map, and grasp angle affordance map. This design allows the network to share latent features essential for grasping while efficiently predicting both grasping types using lightweight head layers with fewer parameters. As a result, the network achieves efficient and simultaneous inference for both suction and finger-grasping tasks. This method addresses the limitation of traditional grasping strategies that require object recognition prior to grasping.

In detail, the multi-action grasp network is defined as $f_{\theta}: I_{rgb}, I_{d} \rightarrow \{A_{s}, A_{f}, A_{a}\}$, where $A_{s}, A_{f}, A_{a}$ are the suction affordance map, finger affordance map, and grasp angle affordance map, respectively. $A_{s}$ and $A_{f}$ have the same dimension of $\mathbb{R}^{3 \times H\times W}$ representing grasp success, grasp failure, background probability. $A_{a}$ has the dimension of $\mathbb{R}^{12 \times H\times W}$, where each channel contains the probability of the grasp angle on each pixel from -90 to 90 degrees with 15-degree intervals. We select the grasp type and grasp point with the priority of suction grasp if the maximum probability of suction grasp is greater than the threshold $\varepsilon_{s}$.

\begin{equation}
    \hat{u} = \argmax_{u\in U} A_{s}(u)
\end{equation}
where $U$ is a set of grasp candidates, and $\hat{u}$ is the selected grasp point in 2D image coordinates. Otherwise, we select the grasp point that has the maximum probability between suction and finger grasp which can be formulated as follows:
\begin{equation}
    \hat{u} = \argmax_{m\in \{s, f\}, u\in U_{m}} A_{m}(u)
    \label{eq:grasp_type}
\end{equation}
here, $U_{m}$ represents a set of grasp candidate indices for each grasp type $m \in \{s, f\}$. Prioritizing suction grasping is based on the observation that it is more stable than finger grasping in bin-picking, as demonstrated by our experiments in Sec.~\ref{sec:experiments}.

A grasp angle $\varphi$ that maximizes the probability of the grasp angle affordance map $A_{a}$ among the 12 channels in the predicted grasp point is selected as the final grasp angle which represents a rotation angle of the end-effector.:
\begin{equation}
    \varphi = \Phi(\argmax{A_{a}(\hat{u})})
    \label{eq:grasp_angle}
\end{equation}
where $\Phi$ is a function that maps the index of the channel from $A_{a}$ to the corresponding grasp angle.

We trained the network using the multi-task loss function which is defined as a weighted sum of the cross-entropy loss for each task.

During training, we utilized the SGD optimizer with a learning rate of 0.001, a momentum of 0.9, and a weight decay of $1\times 10^{-4}$. The network was trained for 200 epochs with a batch size of 16 on two Nvidia RTX 3090 GPUs.

\begin{figure}[t]
    \centering
    \includegraphics[width=\columnwidth]{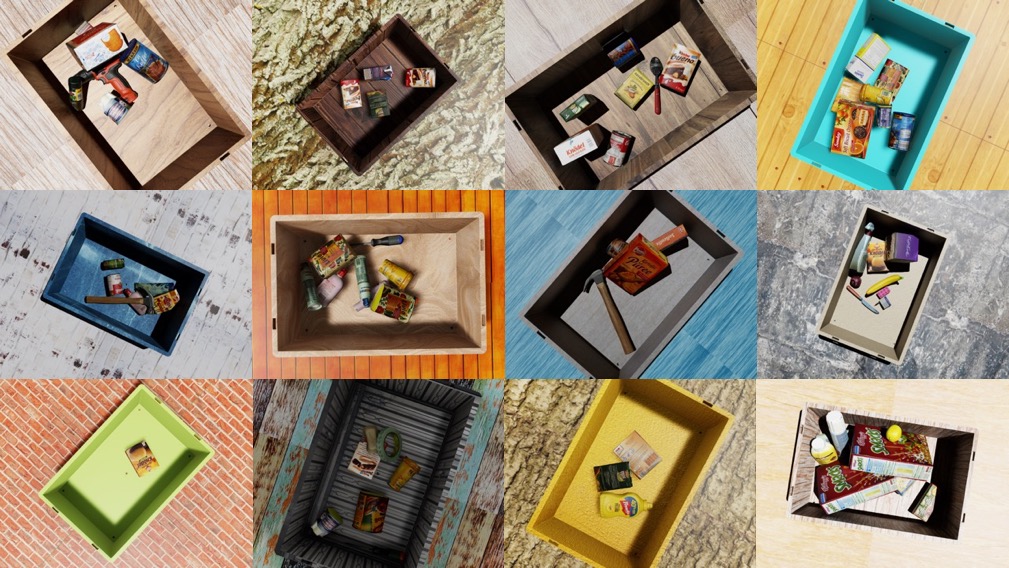}
    \caption{Example images from the synthetic dataset generated using Isaac Sim. The dataset contains 100,000 photo-realistic RGB-D images from 12,500 scenes rendered using Nvidia Isaac Sim. The dataset includes 2.8 billion grasp candidates annotated for both suction and finger grasps}
    \label{fig:synthetic_dataset}
\end{figure}

\subsection{Grasp Dataset Construction and Annotation}

In our previous works~\cite{son2024coasnet,hieu2024enhancing}, we utilized Nvidia Isaac Sim, a photorealistic simulator, to collect image datasets for bin-picking scenarios, ensuring the acquisition of large-scale data. Using this process, we additionally annotated grasp points for each modality by leveraging both simulated image data and privileged information, such as object meshes and object poses. The generated dataset was annotated to satisfy analytic grasping conditions, ensuring consistent labeling across samples. We have also adapted domain randomization to reduce the gap between simulation and the real world. Consequently, the networks trained on this dataset demonstrated stable grasping performance in real-world environments. Fig.~\ref{fig:synthetic_dataset} shows example images from the synthetic dataset generated using Isaac Sim.

\begin{figure}[t]
    \centering
    \includegraphics[width=\columnwidth]{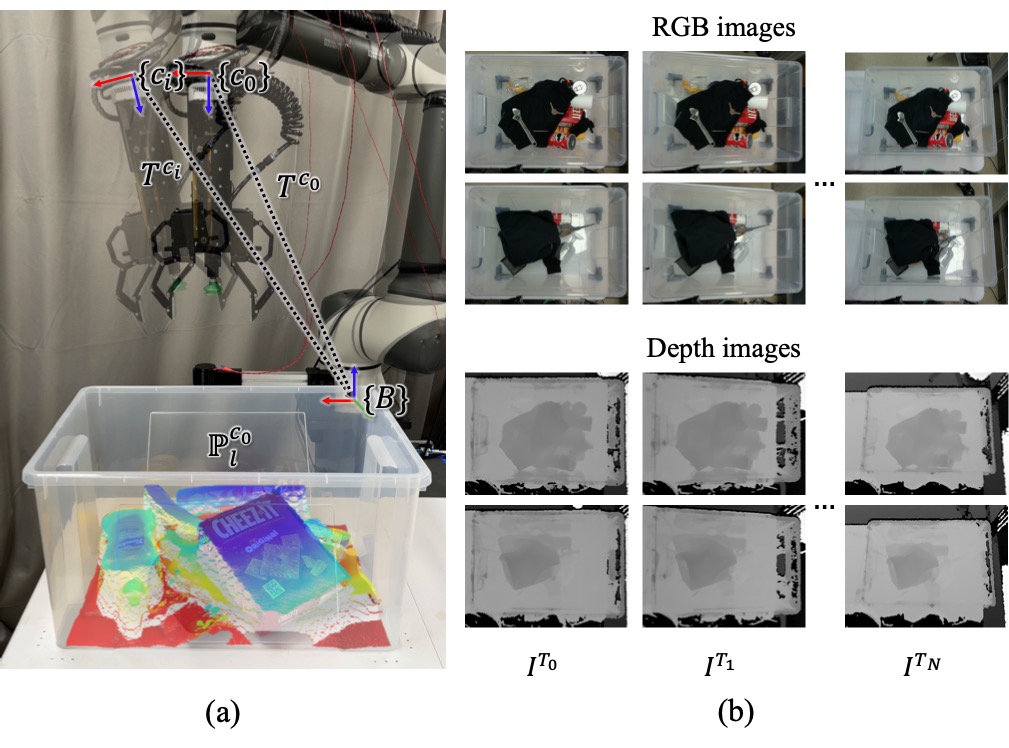}
    \caption{Real-world dataset generation for the multi-action grasp detection network: (a) RGB-D images are captured from a bin-picking scene containing diverse objects, with varying camera poses; (b) Image samples from the dataset captured with different perspectives. Images within each column were captured from the same camera pose and images within each row were captured from the same scene}
    \label{fig:realworld_dataset_generation}
\end{figure}

\begin{figure}[t]
    \centering
    \includegraphics[width=\columnwidth]{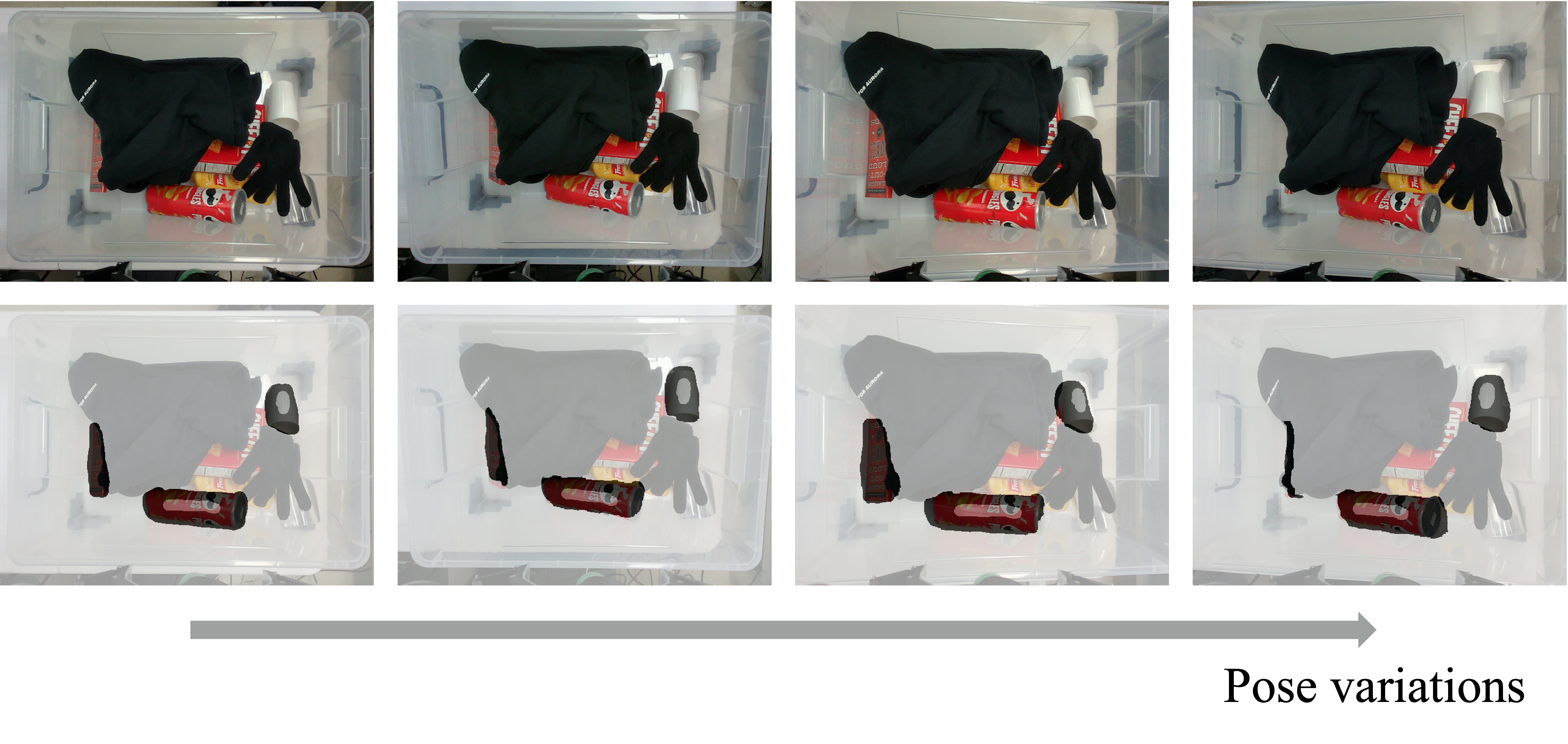}
    \caption{Suction grasping annotation results of the real-world dataset. In the example, images of the same scene are labeled consistently across different camera poses. Labels are transformed to accurately reflect changes in viewpoint, ensuring the annotations correspond to the visible aspects from each camera pose}
    \label{fig:realworld_dataset_annotation}
\end{figure}

In addition to the synthetic dataset, we collected real-world RGB-D images to further enhance the network's generalization capability in real-world scenarios. We utilized spatial information recorded during an image capture process to self-label grasp candidates from a single representative label example, which significantly reduced the labeling cost. In detail, we capture RGB-D images from a bin-picking scene containing diverse objects with a camera attached to the wrist of the manipulator. Then, we vary the camera's position and orientation to capture multiple images of the same scene as shown in Fig.~\ref{fig:realworld_dataset_generation}. Each variation is recorded with the camera's pose relative to its initial posture. We then manually label a single representative image following the same style as the synthetic dataset, which includes three labels: grasp success, grasp failure, and background. After that, the representative scene is converted into a point cloud, with each point assigned a label. Finally, each point is transformed and re-projected into other image planes of different camera poses using the camera's extrinsic parameters to annotate other images within the same scene. This can be formulated as follows:
\begin{equation}
    \mathbb{P}_{l}^{c_{i}} = T^{c_{i}} T^{c_{0}^{-1}}(\mathbb{P}^{c_{0}}_{l}) \quad \forall i \in \{1, 2, ..., N\}
\end{equation}
where $\mathbb{P}_{l}^{c_{i}}$ is the 3D point cloud obtained from pixels corresponding to label $l$ in the $i$-th labeled 2D image, $T^{c_{i}}$ is the camera pose corresponding to the $i$-th image, $T^{c_{0}}$ is the initial camera pose at which the representative image was captured, and $\mathbb{P}^{c_{0}}_{l}$ is the point cloud obtained from the initial pose where the representative image was captured. We set the number of camera pose variations, $N$, to 15, which is sufficient to introduce variability into a dataset.

This process ensures that images remain labeled correctly even when seen from different perspectives. Fig.~\ref{fig:realworld_dataset_annotation} shows the labeling results of the real-world dataset.

In total, the grasp dataset comprises 100,000 synthetic images annotated with 2.8 billion grasp candidates and 3,000 real-world images annotated with 60 million grasp candidates.

\subsection{Collision Avoidance Planning}

In the bin-picking task, avoiding collision is crucial to eliminate the risk of damaging the gripper and the environment. Previous studies have simplified the collision avoidance problem by using a bin with a low height or not using a bin at all~\cite{chen2023efficient, cao2021suction, fang2020graspnet, kumra2020antipodal, mahler2018dex, mahler2017dex}. However, in real-world scenarios, it is general that the bin's height is usually higher than the gripper's height, making it difficult to avoid collisions.

Our collision avoidance planning strategy leverages the gripper's reconfigurable finger motion and reciprocating suction cup to actively avoid collisions with the environment. As introduced in Sec.~\ref{sec:gripper_design}, ReC-Gripper utilizes its functionality to avoid collisions in various scenarios. To leverage this capability, it is essential to identify the predicted grasp candidate's position within the bin and refine it if necessary. For this purpose, we determine the location of the bin containing the target object in the image, identify grasp candidates within the bin, and plan collision-free grasps accordingly.

In detail, we estimate the pose of the bin by segmenting the bin region using the Segment Anything Model (SAM)~\cite{kirillov2023segment} from the RGB observation. From the resulting segmentation mask, we extract the bin boundary and compute a rotated rectangular fit to obtain the bin orientation and a consistent reference point on the bin in image coordinates. Since the camera is fixed in our setup, the distance between the camera and the bin is treated as a known constant. Using the calibrated camera intrinsics and this predefined distance, the selected 2D reference point is back-projected to 3D to recover the bin frame translation in the camera coordinate. Combining this translation with the rectangle-derived orientation defines the bin pose as a transformation matrix with respect to the camera frame, $T_{b}^{c} \in \mathbb{S}\mathbb{E}(3)$.

After defining the pose of the bin, the grasp point $\hat{u}$ is back-projected into the camera frame as $\hat{p^{c}} \in \mathbb{R}^{3}$. This point is then transformed into the bin's coordinate as $\hat{{p}^{b}} = ({T}^{c}_{b})^{-1}\hat{p^{c}}$ and matched with a region that indicates a predefined collision-related region tag. This tag is defined as the distance from the edge of the bin. If the grasp candidate is located in the collision-related region, the candidate is checked for refinement. The region can be classified into two sections: edge region and corner region. The edge region is defined as the area near the edge of the bin, and the corner region is defined as the area near the corner of the bin.

In the case of a suction grasp, if no collisions are expected, The grasp pose $T$ for the grasp candidate is defined using 3D grasp point $\hat{p^{c}}$, the normal vector $n^{c}$ extracted from the point cloud, and gripper rotation angle $\varphi$, which is estimated from the grasp angle affordance map $A_{a}$. The gripper configuration $C$ for the grasp candidate is defined as $C=(s_{out}, f_{open}, f_{dft})$.

If the suction grasp candidate is in the collision-related region and the normal vector is facing the near edge or corner, the grasp candidate is refined to avoid collisions. A grasp pose $T$ is defined from the 3D grasp point $\hat{p^{c}}$ and a normal vector perpendicular to the table. The gripper rotation angle $\varphi$ is set to angle $\varphi_{safe}$ that makes the flat side of the gripper face the edge or corner. The refined gripper configuration $C$ is defined as $C=(s_{out}, f_{open}, f_{rot})$.

\begin{figure}[t]
    \centering
    \includegraphics[width=0.99\columnwidth]{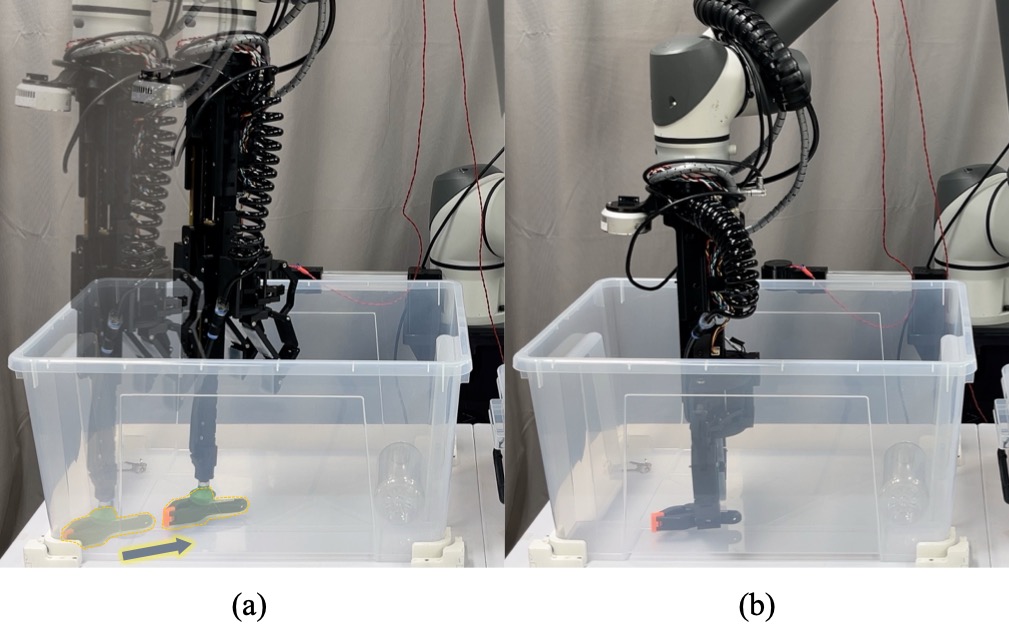}
    \caption{A pushing motion to prevent collisions and secure sufficient space for finger grasping: (a) The gripper extends the suction cup, rotates its fingers, and moves the object toward the center of the bin. (b) The gripper grasps the object with a finger grasp after obtaining sufficient grasping space}
    \label{fig:push_motion_primitive}
\end{figure}

For a finger grasp, the grasp pose $T$ is determined using a top-down approach using the 3D grasp point $\hat{p^{c}}$. The gripper configuration $C$ is defined as $C=(s_{in}, f_{close}, f_{dft})$. If a collision between the fingers and the bin is expected, the grasp is refined. While maintaining the grasp position and rotating the end-effector to $\varphi_{safe}$, the gripper pushes the object toward the center of the bin with the configuration $C=(s_{out}, f_{open}, f_{rot})$. Fig.~\ref{fig:push_motion_primitive} illustrates the pushing motion for a finger collision avoidance strategy.

\subsection{Grasp Fusion}

\begin{figure}[t]
    \centering
    \includegraphics[width=\columnwidth]{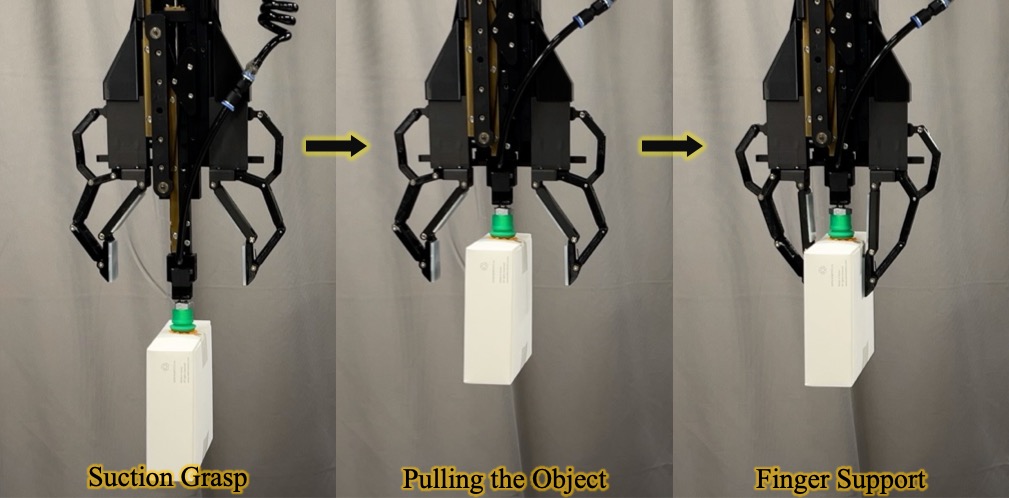}
    \caption{Grasp fusion strategy for enhanced grasp stability. The gripper first grasps the object using the suction cup and then pulls the suction cup back to bring the object to the fingers for additional grasping}
    \label{fig:grasp_fusion}
\end{figure}

Our method can further enhance its grasp stability by fusing both suction and finger grasping. As shown in Fig.~\ref{fig:grasp_fusion}, the fusion strategy involves grasping the object using the suction cup and then pulling the suction cup back to bring the object to the fingers for additional grasping. This seamless transition is facilitated by aligning the gripper's orientation to the predicted finger grasp angle from the grasp angle affordance map $A_{a}$ during the initial suction approach. The grasp configuration for the fusion can be formed as $C=(s_{out}, f_{open}, f_{dft}) \rightarrow (s_{in}, f_{close}, f_{dft})$. This strategy effectively prevents the object from falling during rapid movements after grasping, thereby ensuring stability. In Sec.~\ref{sec:experiments}, our experiments confirmed that this approach significantly enhances grasp stability.

\section{Surface Material Detection Network}
Along with the multi-action grasp detection network, we propose a Surface Material Detection Network (SMD-Net) which discriminates the surface material of the target grasp region to select the appropriate grasp strategy based on the surface material of the target object.

\subsection{Network Architecture}
\begin{figure}[t]
    \centering
    \includegraphics[width=\columnwidth]{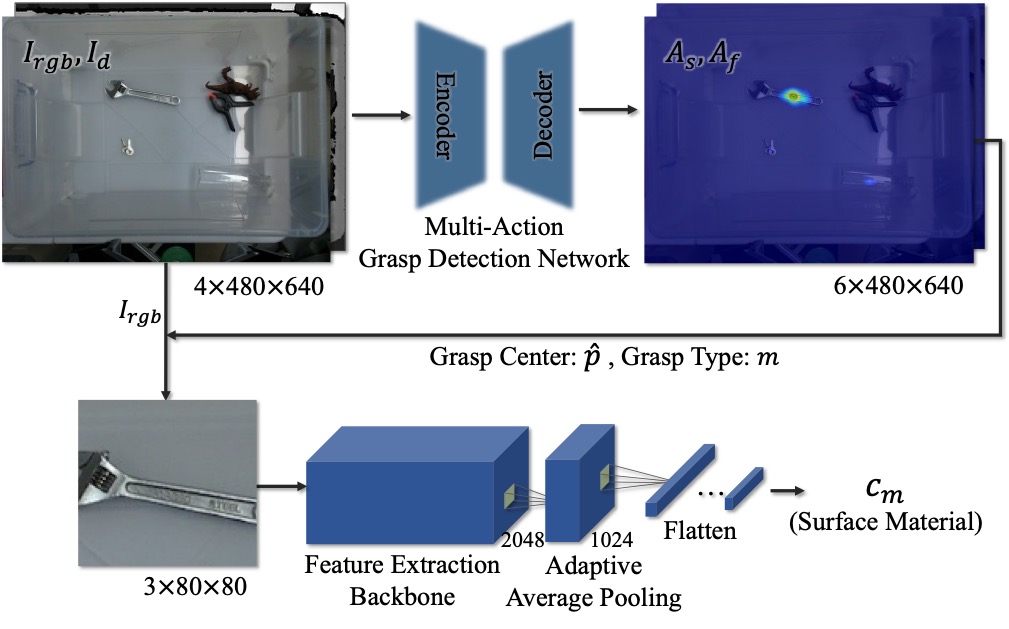}
    \caption{Surface material classification pipeline using SMD-Net. The pipeline takes an RGB image and crops it around the grasp point predicted by the multi-action grasp detection network. The cropped image is then input to the SMD-Net to predict the surface material category}
    \label{fig:SMD-Net_architecture}
\end{figure}

Surface Material Detection Network (SMD-Net) discriminates the surface material of the target grasp region into three categories: metallic, transparent, and others. This network takes an RGB image cropped around the grasp point, resized to  $80 \times 80$, and outputs the surface material category. SMD-Net extracts the feature of the image using a DenseNet-121~\cite{huang2017densenet}, which is then followed by an average pooling layer and a fully connected layer to predict the surface material. Specifically, SMD-Net is defined as $g_{\phi}: \acute{I}_{rgb} \rightarrow c_{m}$, where $\acute{I}_{rgb}$ is a cropped RGB image, $c_{m}$ is a surface material. Fig.~\ref{fig:SMD-Net_architecture} illustrates the surface material prediction pipeline and the network architecture of SMD-Net. This network is designed to provide additional information to the multi-action grasp detection network, enabling it to select the appropriate grasp configuration based on the surface material of the target object and broaden the range of objects that can be grasped. If the surface material is metallic, the network selects the electromagnetic grasping strategy which is described in Sec.~\ref{sec:gripper_design}. If the surface material is transparent, the network selects the suction grasping strategy, in which the robot plans to grasp the object using only the suction cup with slow and gentle motion until the pressure sensor confirms that the suction is sealed. For other surface materials, the network selects the grasp type based on the aforementioned grasp selection strategy. No matter what the surface material is, the collision avoidance strategy is applied identically to prevent collisions with the environment.

\subsection{Surface Material Dataset}
We generated a SMD-dataset containing 9k images of various surface materials, including metallic, transparent, and others. Images were collected from the result of the inference of the multi-action grasp detection network with various grasp scenarios. We randomly sampled points from the grasp region following a gaussian distribution centered at the grasp point. Then RGB images were cropped around the sampled points to have a dimension of $80\times 80$. Finally, the images were annotated with the corresponding surface material labels.

\begin{figure}[t]
    \centering
    \includegraphics[width=0.82\columnwidth]{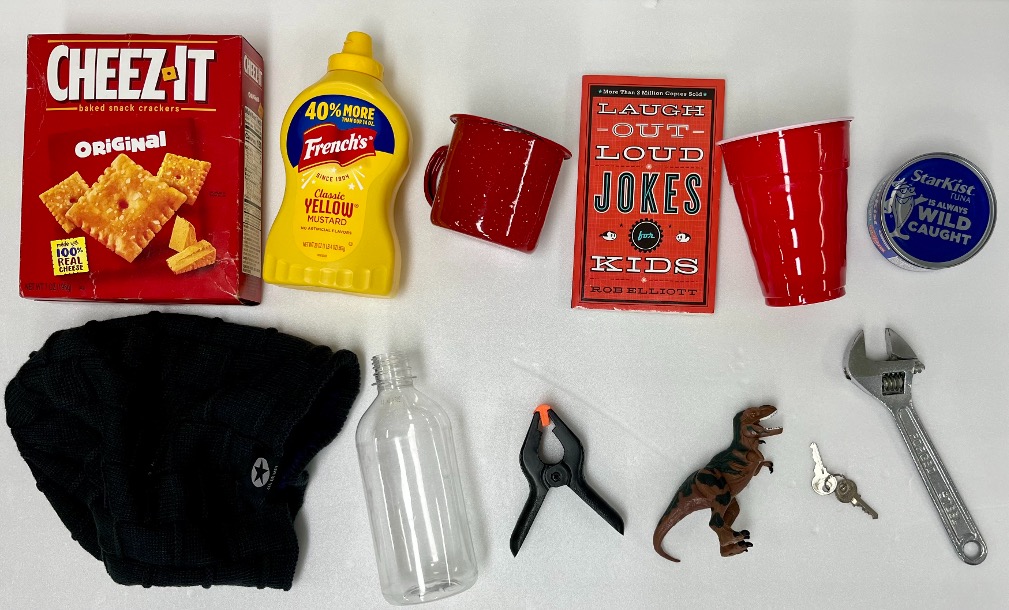}
    \caption{Target objects used in our experiments. The objects vary in shape, size, and properties, including deformable and transparent objects. A subset of these objects is included in the dataset used to train the multi-action grasp detection network, while the remaining objects are used to evaluate its generalization capability}
    \label{fig:target_objects}
\end{figure}

\section{Experiments}\label{sec:experiments}

\begin{figure*}[t]
    \centering
    \includegraphics[width=\textwidth]{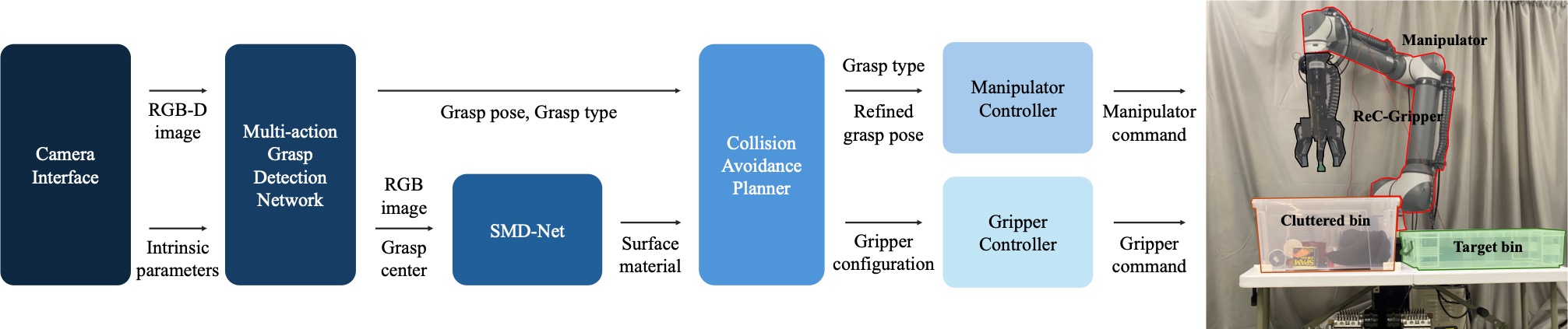}
    \caption{Overview corner-grasp system. The multi-action grasp detection network predicts both suction and finger grasp points from the input RGB-D image, while the surface material detection network classifies the surface material of the target grasp region. Based on the predictions from both networks, the system selects and refines a grasp strategy and executes it using the ReC-Gripper and the manipulator. The system moves objects from a cluttered bin to a target location}
    \label{fig:system_integration}
\end{figure*}

We evaluate the proposed system in real-world cluttered bin-picking scenarios using a Rainbow Robotics RB10 6-DoF robot arm equipped with the ReC-Gripper. A RealSense L515 RGB-D camera is mounted on the robot wrist to capture observations of the scene. We evaluate performance on diverse objects with varying shapes, sizes, and materials under cluttered conditions; the objects used in our experiments are shown in Fig.~\ref{fig:target_objects}. The multi-action grasp detection network is implemented in PyTorch, while robot control and motion planning are integrated through the ROS framework to form an end-to-end corner-grasp system, as illustrated in Fig.~\ref{fig:system_integration}. All experiments are deployed on a workstation with an Intel i7-10700F CPU and an NVIDIA RTX 3060 Ti GPU. Overall, these experiments are designed to validate the effectiveness and robustness of our approach in practical settings.

\begin{table*}[t]
    \centering
    \setlength{\tabcolsep}{2.5pt}
    \caption{Comparison of the Proposed System in Different Tasks and Ablation Methods}
    \label{tab:experiment_results}
    \small
    \begin{tabular}{c c c c c c c c c c}
      \toprule
      \toprule
      \multicolumn{3}{c}{\textbf{Experiment description}} & \multicolumn{2}{c}{$\quad\quad$\textbf{Performance $(\%)$}$\quad\quad$} & \multicolumn{5}{c}{$ \quad\quad$ \textbf{Grasp type $(\%)$}$\quad\quad$} \\
      \cmidrule(lr){4-5} \cmidrule(l){6-10}
      & & & $\quad\quad$SR$\quad\quad$ & $\quad\quad$CR$\quad\quad$ & Suction & Suction$_{ca}$ & Finger & Push & Magnetic\\
      \midrule
      \multirow{2}{*}{(a) Task}
        & \multicolumn{2}{c}{Table-top} & 90.00 & 98.33 & 56.49 & -- & 25.95 & -- & 17.56 \\
        \addlinespace[1pt]
        \cdashline{2-10}[0.5pt/1.5pt]
        \addlinespace[3pt]
        & \multicolumn{2}{c}{Bin} &86.27 & 96.67 & 22.76 & 21.38 & 23.45 & 15.86 & 16.55 \\
      \midrule
      \multirow{2}{*}{(b) Ablation}
        & \multicolumn{2}{c}{No-avd} & 45.70 & 75.00 & 59.78 & -- & 22.83 & -- & 17.39\\
        \addlinespace[1pt]
        \cdashline{2-10}[0.5pt/1.5pt]
        \addlinespace[3pt]
        & \multicolumn{2}{c}{Greedy} & 83.02 & 95.00 & 19.28 & 20.48 & 28.92 & 15.66 & 15.66 \\
      \bottomrule
      \bottomrule
    \end{tabular}
  \end{table*}

\begin{figure*}[t]
    \centering
    \includegraphics[width=\textwidth]{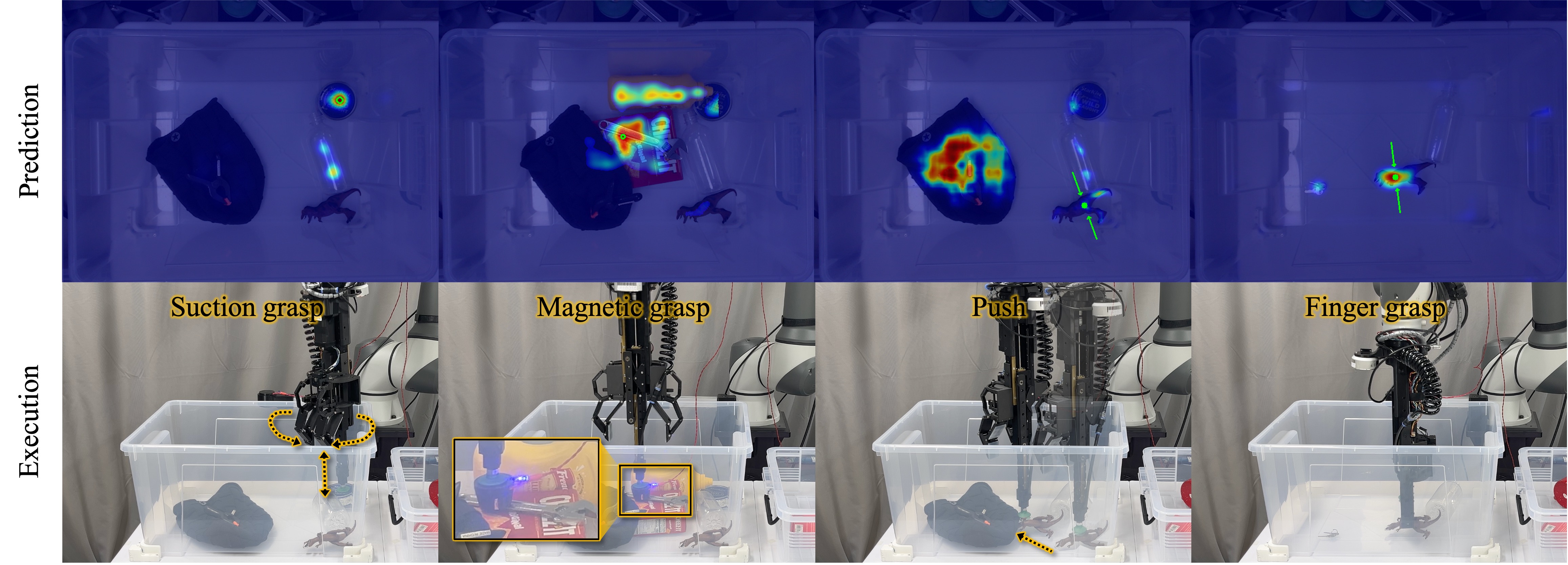}
    \caption{Visualization of grasp selection in a cluttered bin. For each step, the top image shows the real robot execution, and the bottom image visualizes the predicted grasp-affordance heatmap and the selected grasp pose. The suction grasp candidate with the highest probability is indicated by a green circle, and the finger grasp candidate with the highest probability is indicated by a green circle with green arrows representing the grasp angle}
    \label{fig:grasp_squence}
\end{figure*}
\begin{figure}[th!]
    \centering
    \includegraphics[width=\columnwidth]{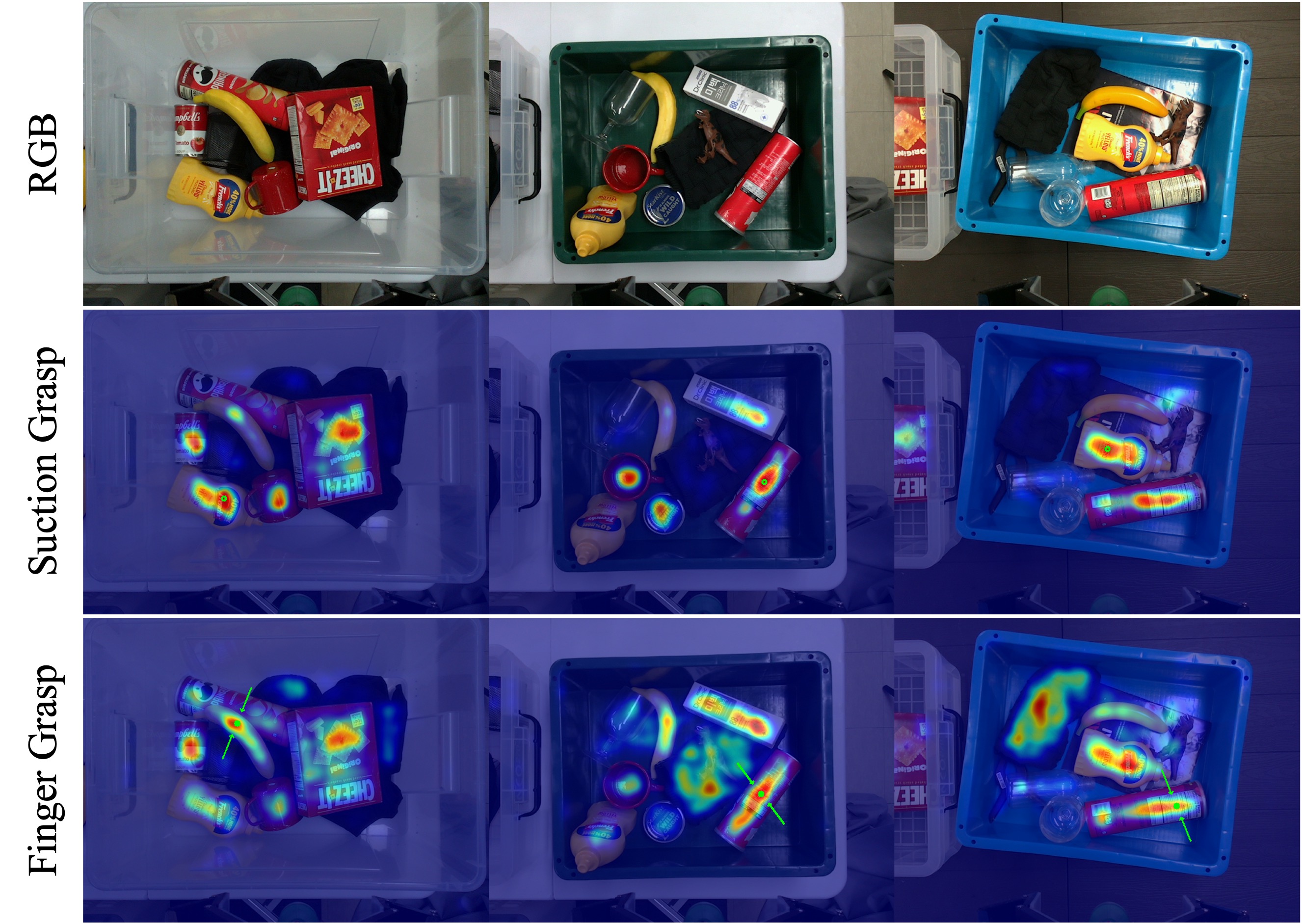}
    \caption{Generalization capability of the multi-action grasp detection network in various scenarios. These scenarios include variations in background, bin configuration, and objects. The network identifies suitable grasp candidates for both suction and finger grasping in each scenario}
    \label{fig:generalization}
\end{figure}

\subsection{Grasping in Cluttered Scenarios}

\begin{figure}[t]
    \centering
    \includegraphics[width=\columnwidth]{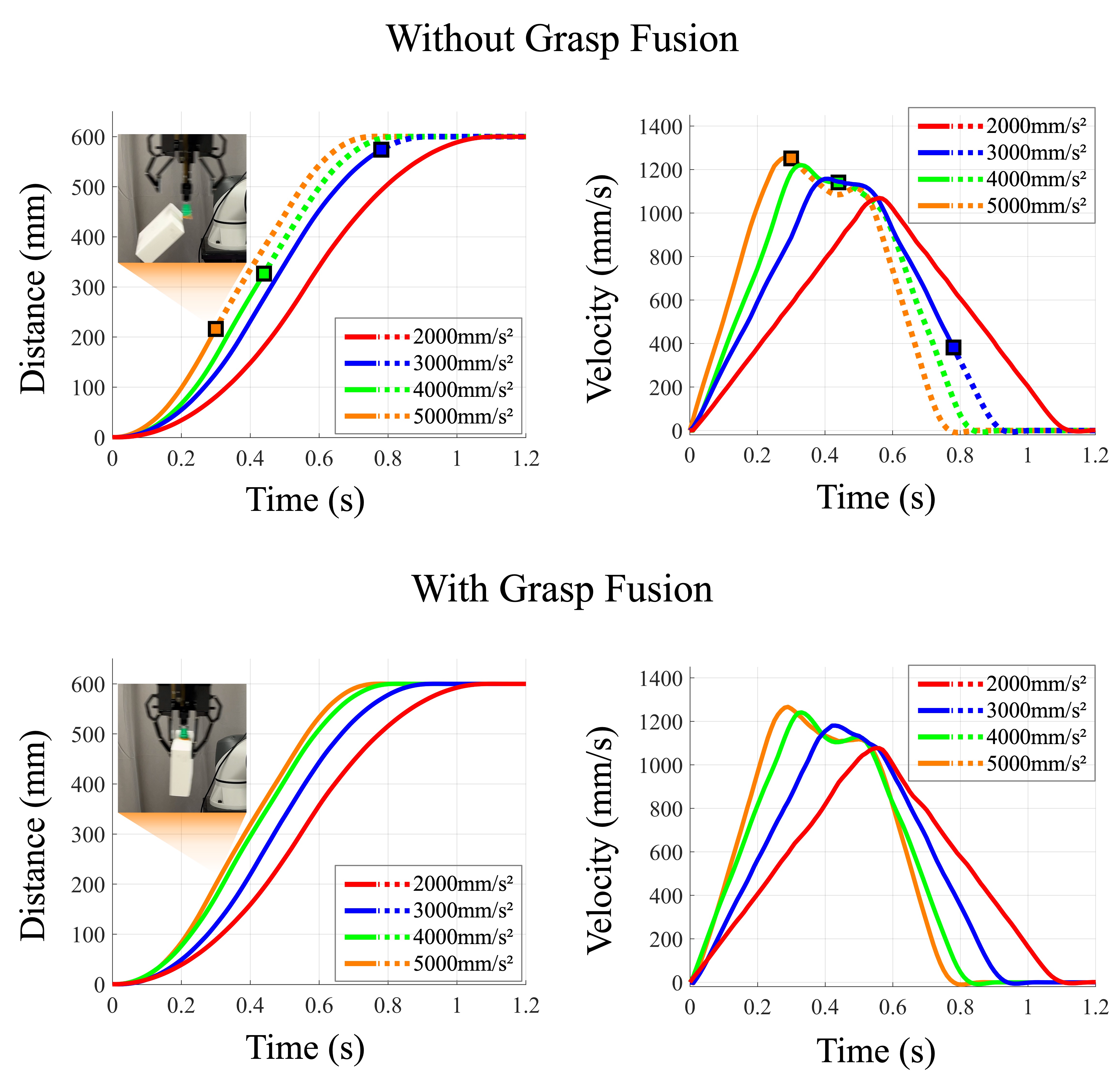}
    \caption{Experimental results of the grasp fusion strategy. We show both trajectories and velocities of the gripper during experiments. The solid line represents that the gripper is in contact with the object, and the dashed line represents that the gripper has lost contact with the object}
    \label{fig:grasp_fusion_res}
\end{figure}
First, to evaluate the picking performance of our proposed system in cluttered environments, we conduct real-world experiments using both table-top and bin-picking setups. In the table-top picking task (\textbf{Table-top}), we disable the collision avoidance strategy since the gripper can safely grasp objects with minimal collision risk. In contrast, the bin-picking task (\textbf{Bin}) uses a bin with a depth of 25 cm, specifically chosen to evaluate the effectiveness of our collision avoidance strategy.

We perform 10 experiments for each task, measuring the system’s success rate and clear rate. In the following tables, we use SR (success rate) and CR (clear rate) to denote the proportion of successful grasps and the proportion of successfully grasped objects among the objects placed in the workspace, respectively. We also report the proportion of grasp types selected by the system in each scenario, as detailed in Table~\ref{tab:experiment_results}. The grasp types listed in the table are defined as follows: suction grasp (Suction), suction grasp with collision avoidance (Suction$_{ca}$), finger grasp (Finger), pushing motion (Push), and electromagnetic grasp (Magnetic). For methods that do not incorporate a collision avoidance strategy, data for Suction$_{ca}$ and Push are not reported.

As shown in Table~\ref{tab:experiment_results}, our proposed system achieves high success and clear rates across both the table-top and bin-picking scenarios. The grasp selection capability effectively adapts to each environment, confirming the method’s effectiveness. Although performance slightly decreased during bin-picking due to the increased complexity, the collision avoidance strategy significantly improved the system’s ability to prevent collisions, enabling stable and effective grasping in cluttered environments. To better show the effectiveness of the collision avoidance strategy, we experiment with a baseline system that does not use collision avoidance (\textbf{No-avd}). During the experiment, we manually remove an object from the bin if the system fails to grasp it due to collisions for three consecutive attempts and apply a penalty to the clear rate. The results show that the proposed collision avoidance strategy significantly improves the system’s performance, as evidenced by the increased success rate and clear rate compared to the baseline system.

We also compare our grasp selection strategy with a greedy grasp selection strategy (\textbf{Greedy}), which selects the grasp type with the highest probability as defined in Eq.~\ref{eq:grasp_type}. The results show that our grasp selection strategy effectively clears cluttered environments by prioritizing suction grasping, which offers greater stability compared to finger grasping in bin-picking scenarios. The slightly lower success rate observed for the greedy method compared to our proposed strategy is primarily due to failures in finger grasping within cluttered environments at an initial phase of the task, where finding collision-free grasp points is particularly challenging.

We also visualize snapshots of the grasp sequence in a bin-picking scenario in Fig.~\ref{fig:grasp_squence}. The figure illustrates the predicted grasp candidates and the selected grasp pose at each step. The system effectively selects appropriate grasp types based on the predicted affordance maps, demonstrating its capability to adapt to different scenarios within a cluttered environment.

In addition, we evaluate whether our system robustly detects grasp points across various environments. To achieve this, we conduct experiments by varying the background, bin properties, and objects. Specifically, we change the background’s color and texture and use bins with different colors, sizes, and poses. Objects are also varied in shape and size. As shown in Fig.~\ref{fig:generalization}, our system successfully identifies suitable grasp candidates in these varied experimental conditions. These results confirm that our system generalizes effectively across different environments and conditions.

\subsection{Grasp Fusion Evaluation}
We further investigate the effectiveness of our grasp fusion strategy by conducting comparative experiments. Specifically, we evaluate the performance of the robot to maintain grasp stability while moving an object rapidly to a target location under varying accelerations. The robot either uses suction grasping or fusion grasping strategy to grasp and move the object. During these experiments, we vary the robot acceleration in task space within a range from 2,000mm/s$^2$ to 5,000mm/s$^2$. As shown in Fig.~\ref{fig:grasp_fusion_res}, both strategies can resist the wrenches at a low acceleration. However, as the acceleration increases, the suction cup is required to withstand significant wrenches caused by inertial forces, which leads to a failure to maintain grasp stability. In contrast, the fusion grasping strategy effectively prevents objects from falling during fast movements. This result demonstrates that grasp fusion effectively enhances stable grasping in dynamic scenarios.

\subsection{Effectiveness of Dataset}
\begin{table}[t]
    \centering
    \setlength{\tabcolsep}{6pt}
    \caption{Comparative Analysis of Grasp Performance Using a Multi-Action Grasp Detection Network Trained on Different Datasets}
    \label{tab:dataset_ablation}
    \begin{tabular}{c c c c c c}
        \toprule
        \multirow{2}{*}{$\quad$\textbf{Method}}
        & & \multicolumn{2}{c}{\textbf{In-domain}} & \multicolumn{2}{c}{\textbf{Out-of-domain}} \\
        \addlinespace[1pt]
        & & SR & CR & SR & CR\\
        \midrule
        \multicolumn{2}{c}{Real-full} & 70.00 & 91.67 & 61.18 & 84.72\\
        \addlinespace[1pt]
        \addlinespace[1.5pt]
        \multicolumn{2}{c}{Synth-sub} &54.04 & 79.17 & 55.38 & 81.67\\
        \addlinespace[1.5pt]
        \multicolumn{2}{c}{Synth-full} &65.78 & 88.33 & 65.71 & 85.00\\
        \addlinespace[1.5pt]
        \multicolumn{2}{c}{Synth-real-full} &\textbf{86.27} & \textbf{96.67} & \textbf{85.62} & \textbf{95.83}\\
        \bottomrule
    \end{tabular}
\end{table}

To validate the influence of the dataset on the performance of the multi-action grasp detection network, we conduct experiments using different datasets. We compare the performance of the system using multi-action grasp detection networks trained on various dataset combinations: \textbf{Real-full} represents a model trained only on the real-world dataset, \textbf{Synth-sub} is a model trained on the subset of our synthetic dataset matched in size to the real-world dataset, \textbf{Synth-full} is a model trained on the full synthetic dataset which consists of 100,000 RGB-D images, and \textbf{Synth-real-full} a model collaboratively trained on the full synthetic dataset with and the real-world dataset. The models are evaluated in terms of success rate and clear rate in both In-domain and Out-of-domain. The In-domain scenario is defined as the same scene where the real-world dataset was collected, while the Out-of-domain is defined as different scenes from the real-world dataset. The results are shown in Table~\ref{tab:dataset_ablation}. In the in-domain evaluation setting, the model trained exclusively on the real-world dataset (Real-full) outperforms the models trained solely on synthetic datasets (Synth-sub and Synth-full), regarding success rate and clear rate. This performance gap primarily arises from the limited generalization capability of synthetic-only model when tasked with grasping objects possessing physical characteristics challenging to accurately simulate. Notably, the Synth-sub model exhibits a marked performance degradation in both success and clear rates, underscoring the critical role that dataset scale plays in enhancing grasping capabilities derived from synthetic data. In contrast, results from the Out-of-domain evaluation setting show that the performance of the Real-full drops significantly, while the performance of the Synth-full and Synth-real-full models remains relatively stable. This highlights its limited generalization beyond its original training distribution. Notably, our proposed model (Synth-real-full), leveraging collaborative training with synthetic and real-world data, achieves superior performance across both in-domain and out-of-domain evaluations. These demonstrate the efficacy and robustness of our collaborative training approach, combining the advantages of real-world specificity with the generalizability provided by synthetic data augmentation.

\subsection{Comparison with Other Methods}

\begin{figure}[t]
    \centering
    \includegraphics[width=\columnwidth]{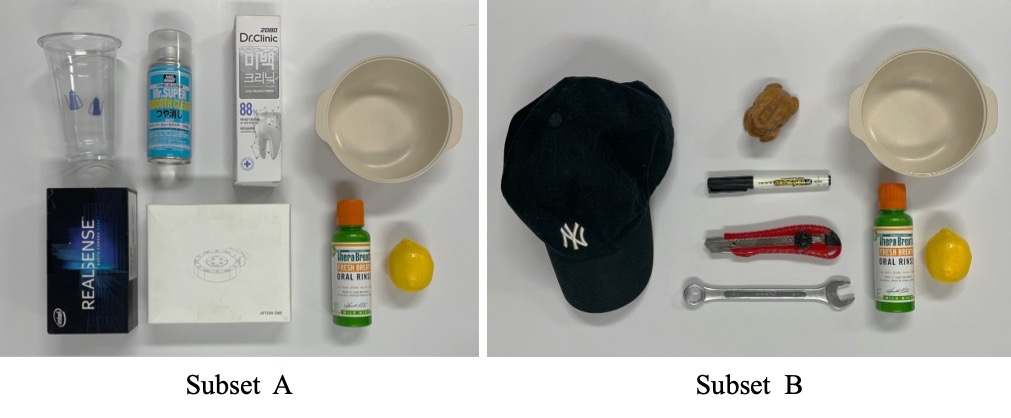}
    \caption{Target object subsets used in our experiments. Each subset has 8 objects, with 3 objects overlapping between the two subsets. All objects are unknown to the multi-action grasp detection network}
    \label{fig:target_objects_AB}
\end{figure}

To compare with existing approaches, we evaluate suction and finger grasping separately in a table-top picking task using representative prior methods for each grasp mode. As baselines are designed for only one mode, we prepare two evaluation subsets that match their intended use. The suction evaluation uses an object subset that can be reliably picked with suction, and the finger-grasp evaluation uses another object subset that is well suited for finger grasping. Each subset contains 8 objects, with 3 objects overlapping between the two subsets. Fig.~\ref{fig:target_objects_AB} shows the objects in each subset. All objects in both subsets are unknown to the multi-action grasp detection network, ensuring a fair evaluation of generalization capability.

\begin{table}[t]
    \centering
    \setlength{\tabcolsep}{8pt}
    \caption{Comparison with Prior Single-mode Grasp Detection Methods on the Table-Top Picking Task}
    \label{tab:comparison_single_mode}
    \begin{tabular}{cccc}
    \toprule
    \addlinespace[2pt]
    \textbf{Method}                 & \textbf{Objects}  & \multicolumn{1}{l}{\textbf{SR}} & \multicolumn{1}{l}{\textbf{CR}} \\ \hline
    \addlinespace[2pt]

    \cite{cao2021suction} & Subset A & 64.00                            & 80.00                              \\
    \cite{zeng2022robotic}          & Subset A & 80.95                            & 85.00                              \\
    Ours (suction) & Subset A & \textbf{88.10}                            & \textbf{92.50}                              \\
    \addlinespace[2pt]

    \hline
    \addlinespace[2pt]

    \cite{gou2021rgb}            & Subset B & 60.78    & 77.50                            \\
    \cite{fang2020graspnet}     & Subset B & 67.35                           & 82.50                              \\
    Ours (finger)  & Subset B & \textbf{73.91}                            & \textbf{85.00}                              \\

    \bottomrule

    \end{tabular}
    \end{table}
For our method, we additionally report single-mode performance by running the same network and executing grasps using only the affordance map of the corresponding mode. This allows us to compare fairly against single-mode baselines while also demonstrating that our unified model can be deployed either as a multi-action predictor or as a single-action grasp detector.

Table~\ref{tab:comparison_single_mode} summarizes the results of these evaluations. In the suction-only setting on Subset A, our method consistently outperforms suction-specific baselines. In the finger-only setting with Subset B, our approach also achieves higher grasp success and clear rates than prior finger-grasp detectors. These results demonstrate that our network not only predicts both suction and finger grasp affordances within a single model, but can also be deployed effectively in a single-mode manner by using only the corresponding output, while maintaining competitive performance against mode-specific methods. We hypothesize that the performance of our method stems from training on large-scale and diverse synthetic data augmented with additional real-world data, which improves generalization to unseen objects and scenarios.

We further evaluate a practical scenario by combining the objects from both mode-specific subsets into a table-top picking task. This mixed evaluation includes 13 objects collected from the two previous experiments, resulting in a more diverse set in terms of shape, size, and surface properties. Table~\ref{tab:comparison_mixed} reports the results. In this setting, single-mode approaches are inherently constrained because a fixed grasp modality cannot robustly cover all object types, and performance degrades when the object set contains cases that are unfavorable to the chosen mode. In particular, the finger-only baseline performs better than the suction-only baseline in this mixed setting, since finger grasping can generally handle a wider range of object geometries and materials than suction. Nevertheless, a single grasp modality still exhibits clear limitations when faced with the full spectrum of objects in clutter. Our full multi-action system achieves the best overall performance, demonstrating the benefit of jointly predicting and selecting between suction and finger grasps according to the object and scene conditions. Also, as we designed the system to jointly predict both grasp types from a shared feature representation, the inference time is only slightly higher than that of single-mode baselines, while providing the flexibility to choose the most suitable grasp type for each object. Moreover, removing SMD-Net leads to a noticeable drop in performance, indicating that material-aware reasoning, together with auxiliary grasping strategies, plays an important role in improving robustness across diverse objects while adding only negligible inference overhead.

\begin{table}[t]
    \centering
    \setlength{\tabcolsep}{7pt}
    \caption{Comparison in a mixed-object bin-picking task with 13 objects formed by combining the evaluation subsets}
    \label{tab:comparison_mixed}
    \begin{tabular}{cccc}
    \toprule
    \addlinespace[2pt]
    \textbf{Method}                   & \textbf{SR} & \textbf{CR} & \textbf{Time (ms)}\\ \hline
    \addlinespace[2pt]

    \cite{zeng2022robotic}
              & 35.87 & 50.77   & 66     \\

    \cite{fang2020graspnet}
         & 55.06 & 75.38   & 224     \\
    Ours (w/o SMD)     & 83.19    & 92.31   & 81     \\
    Ours     & \textbf{88.89} & \textbf{98.46}   & 83     \\

    \bottomrule

    \end{tabular}
    \end{table}

\section{Conclusion}

This paper presents Corner-Grasp, a robotic bin-picking system designed for cluttered and constrained environments. We combine a compact suction-finger gripper with a neural network that simultaneously predicts both grasp types from a single RGB-D image. To improve success rates in deep bins, we introduce an active gripper adaptation and collision avoidance strategy, supplemented by material-aware and electromagnetic grasping capabilities. Trained on a hybrid dataset of synthetic and real-world data for enhanced robustness, the system demonstrated reliable grasping and collision prevention in physical experiments. Future work will explore reinforcement learning-based optimal grasp selection and generalized grasp fusion strategies.

\section*{Acknowledgments}
This work was supported in part by the Korea Institute for Advancement of Technology (KIAT) under Grant P0026249 and funded by the Ministry of Trade, Industry \& Energy (MOTIE).

\bibliography{reference}

\end{document}